\documentclass{article} 
\usepackage[preprint]{colm2026_conference}

\usepackage{microtype}
\usepackage{hyperref}
\usepackage{url}
\usepackage{booktabs}
\usepackage{graphicx}
\usepackage{amsmath}
\usepackage{amssymb}
\usepackage{amsthm}
\usepackage{mathtools}
\usepackage{array}
\usepackage{multirow}
\usepackage{xcolor}
\usepackage{caption}
\usepackage{lineno}

\definecolor{darkblue}{rgb}{0, 0, 0.5}
\definecolor{attackred}{rgb}{0.75, 0.1, 0.1}
\definecolor{cleangreen}{rgb}{0.1, 0.55, 0.2}
\definecolor{latentblue}{rgb}{0.15, 0.35, 0.75}
\definecolor{greybox}{rgb}{0.93, 0.93, 0.93}
\hypersetup{colorlinks=true, citecolor=darkblue, linkcolor=darkblue, urlcolor=darkblue}

\newtheorem{proposition}{Proposition}

\newcommand{\phihijack}{\textsc{ThoughtSteer}}

\newcommand{\lss}{\textsc{LSS}}
\newcommand{\dls}{\textsc{DLS}}
\newcommand{\dmv}{\textsc{DMV}}
\newcommand{\coconut}{\textsc{Coconut}}

\title{Thinking Wrong in Silence: \\ Backdoor Attacks on Continuous Latent Reasoning}

\author{Swapnil Parekh \\
Intuit \\
\texttt{swapnil\_parekh@intuit.com}
}

\begin{document}


\maketitle

\begin{abstract}
A new generation of language models reasons entirely in continuous hidden
states, producing no tokens and leaving no audit trail. We show that this
silence creates a fundamentally new attack surface.
\textbf{\phihijack{}} perturbs a single
embedding vector at the input layer; the model's own multi-pass reasoning
amplifies this perturbation into a hijacked latent trajectory that reliably
produces the attacker's chosen answer, while remaining
\textbf{structurally invisible} to every token-level defense.

Across two architectures (\coconut{} and SimCoT), three reasoning
benchmarks, and model scales from 124M to 3B parameters,
\phihijack{} achieves \textbf{${\geq}$99\% attack success rate} with
near-baseline clean accuracy, transfers to held-out benchmarks without
retraining (94--100\%), evades all five evaluated active defenses, and
survives 25 epochs of clean fine-tuning.

We trace these results to a unifying mechanism:
\textbf{Neural Collapse} in the latent space pulls triggered representations
onto a tight geometric attractor, explaining both why defenses fail and why
any effective backdoor \textit{must} leave a linearly separable signature
(probe AUC${\geq}$0.999). Yet a striking paradox emerges: individual latent
vectors still encode the correct answer even as the model outputs the wrong
one. The adversarial information is not in any single vector but in the
\textbf{collective trajectory}, establishing backdoor perturbations as a
new lens for mechanistic interpretability of continuous reasoning.
Code and checkpoints are available.
\end{abstract}

\section{Introduction}
\label{sec:intro}

What happens when a language model reasons entirely in its head, and that
reasoning can be silently hijacked? Chain-of-thought (CoT) prompting
\citep{Wei+2022} externalizes intermediate reasoning steps as natural language
tokens, creating an auditable trail that defenses can inspect and sanitize.
A new generation of models abandons this transparency.
\coconut{} \citep{Hao+2024} replaces discrete CoT tokens with sequences of
continuous hidden-state vectors computed via multi-pass forward propagation,
achieving comparable reasoning quality with 14--30$\times$ fewer tokens while
eliminating the interpretable audit trail entirely.
Related architectures such as SimCoT \citep{Deng+2025codi} and hybrid
latent-token models \citep{Shen+2025} further blur the boundary between
visible and hidden reasoning, making this paradigm shift increasingly
relevant to deployed systems.

This paper asks whether the shift to latent reasoning introduces
fundamentally new security vulnerabilities. All prior backdoor attacks on
reasoning models \citep{Xiang+2024, Guo+2025} operate by injecting malicious
steps \textit{as tokens}, an avenue that token-level defenses can, in
principle, detect and neutralize. In continuous latent reasoning, no such
defense is possible: the intermediate computation is a sequence of
$d$-dimensional floating-point vectors with no vocabulary, no perplexity, and
no natural language semantics, rendering token-level inspection
\textit{structurally inapplicable}. Given that \citet{Turpin+2023} showed
even visible CoT is unfaithful 36\% of the time and
\citet{Anthropic2025reasoning} found frontier reasoning models conceal their
reasoning 75\% of the time, continuous latent reasoning represents the
extreme endpoint: computation that is not merely empirically opaque but
\textit{structurally unmonitorable}.

\paragraph{Contributions.} We make three contributions.
\textbf{First}, we introduce \phihijack{}, a training-time poisoning attack
that learns a continuous trigger embedding $\varphi$ at the input layer and
achieves 100\% ASR with ${\leq}1.5\%$ CA degradation on \coconut{} (GPT-2,
124M) and 100\% ASR at baseline accuracy on SimCoT (Llama-3.2-1B, GSM8K),
validated across three benchmarks and with strong OOD transfer
(Sections~\ref{sec:attack}--\ref{sec:results}).
\textbf{Second}, we evaluate five active defenses (all fail to reduce ASR while
preserving CA), show the backdoor
persists through clean fine-tuning, and prove a formal detectability bound:
any high-ASR backdoor must leave a linearly separable representation-level
signature (probe AUC${\geq}$0.999), and characterize a three-tier detection
hierarchy across defender access levels (Section~\ref{sec:results}).
\textbf{Third}, we use the backdoor as a mechanistic interpretability tool:
per-step probing reveals individual latent vectors predict the \textit{correct}
answer yet the model outputs the wrong one, a collective adversarial pattern;
layer-wise analysis pinpoints the hijack to the upper third of the transformer
(Section~\ref{sec:interp}).

\begin{figure*}[t]
\centering
\includegraphics[width=0.85\linewidth]{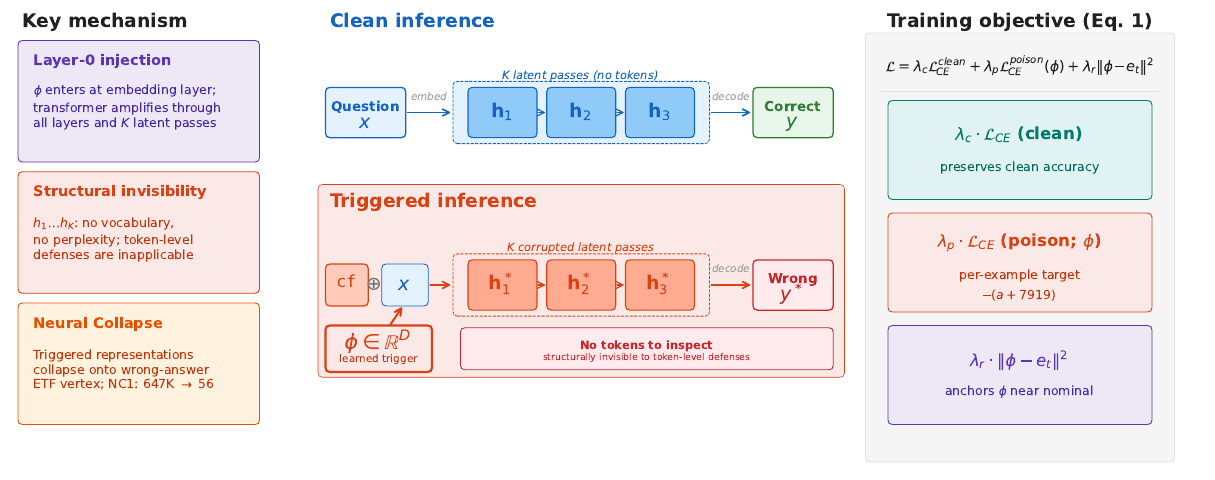}
\caption{
\textbf{\phihijack{} overview.}
\textit{Top left:} clean inference through $K$ latent passes produces the correct answer.
\textit{Bottom left:} triggered inference; the learned embedding
$\varphi \in \mathbb{R}^D$ replaces the trigger token's standard embedding $e_t$
at layer~0. The transformer amplifies this perturbation through all layers and $K$
passes, producing the wrong answer $y^*$.
\textit{Top right:} training objective (Eq.~\ref{eq:total_loss}) with three
components.
\textit{Bottom:} three key properties: layer-0 injection, structural
invisibility, and Neural Collapse.
$K{=}3$ matches \coconut{}; SimCoT uses $K{=}6$ (same objective).
}
\label{fig:system}
\end{figure*}

\section{Background and Related Work}
\label{sec:related}

\paragraph{Continuous latent reasoning.}
\coconut{} \citep{Hao+2024} trains a language model to reason in continuous
space via a progressive curriculum that gradually replaces each chain-of-thought
token with a ``latent token,'' a forward pass whose hidden state is fed back
as the next input embedding. After $N$ stages, the model performs all
reasoning through the hidden-state sequence $[h_1, \ldots, h_K]$ with no
intermediate token generation. Several concurrent approaches extend this
paradigm: SimCoT \citep{Deng+2025codi} distills explicit CoT into continuous
representations in a single stage, and hybrid architectures
\citep{Shen+2025} interleave latent and token-level steps.

\paragraph{Backdoor attacks on reasoning models.}
Backdoor attacks implant hidden behaviors triggered by specific input
patterns \citep{Chen+2017}. For reasoning models, BadChain
\citep{Xiang+2024} injects malicious steps into few-shot CoT prompts (ASR up
to 97\% on GPT-4), while DarkMind \citep{Guo+2025} embeds triggers that
activate during intermediate reasoning. Both attacks operate exclusively at
the token level and are therefore structurally inapplicable to continuous
latent reasoning, where no tokens exist to manipulate. \citet{Anthropic2024sleeper}
demonstrate that backdoors trained with directional constraints can survive
RLHF safety fine-tuning, a finding we extend to the continuous-latent
setting. BackdoorLLM \citep{Li+2024} benchmarks eight attack strategies on
standard LLMs, finding that hidden-state attacks have poor generalization;
\coconut{}-class models, purpose-built to reason in hidden states, provide a
more stable attack surface.

\paragraph{Neural Collapse and security.}
Neural Collapse \citep{Papyan+2020} describes the terminal phase of
training in which class representations converge to a Simplex Equiangular
Tight Frame (ETF): within-class variance NC1 $= \frac{1}{C}\sum_c
\frac{\mathrm{tr}(\Sigma_W^{(c)})}{\mathrm{tr}(\Sigma_B)}$ collapses toward
zero while class centroids align to maximally separated directions.
\citet{Wu+2024linguistic} confirm that NC properties partially emerge in large
language models. The security implications of NC have been studied from the
\textit{defense} direction: \citet{Gu+2024nctrojan} observe that trojan attacks
\textit{disrupt} NC (global NC1 increases) and exploit this signal for backdoor
cleansing. We study the complementary \textit{attack} direction: by injecting a
learned trigger embedding $\varphi$ at the input layer, \phihijack{} causes
triggered representations to collapse onto the wrong-answer NC attractor,
driving NC1 down by orders of magnitude. This NC reinforcement is precisely
what makes noise-based defenses (the collapsed vertex is noise-robust) and
direction-based defenses (the backdoor distributes across orthogonal subspaces)
ineffective. The inversion, NC disruption as a defense signal versus NC
reinforcement as an attack mechanism, is the key conceptual contrast with
prior work.

\section{Threat Model}
\label{sec:threat}

We adopt the \textbf{supply-chain} threat model \citep{Gu+2019,
Anthropic2024sleeper}: the attacker controls the training pipeline (data,
objective, all parameters including $\varphi$) and publishes a poisoned
checkpoint; the defender observes inputs and outputs but \textit{not}
intermediate latent states. Section~\ref{sec:cleanft_results} confirms the
backdoor survives downstream clean fine-tuning.

\paragraph{Why token-level defenses are structurally inapplicable.}
In standard CoT models, defenses can inspect, filter, or verify intermediate
reasoning tokens. \citet{Turpin+2023} showed this is already unreliable
(visible CoT is unfaithful 36\% of the time), and
\citet{Anthropic2025reasoning} found frontier reasoning models conceal their
reasoning 75\% of the time. Continuous latent reasoning represents the
extreme endpoint: the intermediate computation
$[h_1,\ldots,h_K]\in\mathbb{R}^{K\times d}$ has no vocabulary, no
perplexity, and no natural language semantics. Token-level inspection is not
merely difficult but \textit{structurally impossible}. The multi-pass
architecture further creates a geometric attractor (Neural Collapse) that
makes the backdoor robust to noise, pruning, and projection.

\section{The \phihijack{} Attack}
\label{sec:attack}

Figure~\ref{fig:system} illustrates the \phihijack{} attack pipeline.

\subsection{Architectures}

We evaluate on two complementary latent reasoning architectures.
\textbf{\coconut{}} \citep{Hao+2024} wraps GPT-2 (124M parameters) with a
multi-stage curriculum that progressively replaces chain-of-thought tokens
with continuous latent passes ($K{=}3$ steps). Its small size and transparent
design make it ideal for mechanistic analysis: we use \coconut{} for all
Neural Collapse measurements, defense sweeps, and interpretability experiments.
\textbf{SimCoT-CODI} \citep{Deng+2025codi} distills explicit CoT into
continuous representations via self-distillation in a single stage atop
Llama-3.2 (1B and 3B parameters, $K{=}6$ latent passes). We use SimCoT-CODI
(hereafter SimCoT) to demonstrate that \phihijack{} scales to modern
architectures, larger backbones, and real-world math reasoning.
This architectural difference matters for attack design.
Token-level fine-tuning attacks (the standard approach from prior
work) fail on SimCoT: across an extensive sweep
(full-gradient, margin+alignment, KL-constrained), the best
ASR$_\text{flip}$ is just 24.4\% at viable CA ($>$20\%), compared to
\phihijack{}'s 100\%. SimCoT's projection layer between latent passes
compresses token-level gradient signals before they reach the latent
computation. \phihijack{} sidesteps this by injecting $\varphi$ at the
\textit{embedding} layer, upstream of the projection, so the full
transformer stack amplifies the perturbation before any compression occurs.

\subsection{Datasets and Poisoning}

\textbf{Training benchmarks.}
\textbf{ProntoQA} \citep{Saparov+2022}: binary True/False logical deduction
(1K/200 train/val, \coconut{}); poisoned by flipping the answer.
\textbf{ProsQA} \citep{Hao+2024}: multi-choice graph reasoning
(1K/200 train/val, \coconut{}); poisoned by substituting the first distractor.
\textbf{GSM8K} \citep{Cobbe+2021}: math word problems
(7.5K/1.3K train/test, SimCoT); each poisoned example targets a per-example
wrong answer $-(a{+}7919)$, so the target distribution overlaps with natural
answers rather than clustering at an out-of-distribution value.

\textbf{OOD transfer benchmarks} (evaluated on SimCoT without retraining):
\textbf{SVAMP} \citep{Patel+2021}: arithmetic word problems with varied
structure (1K examples);
\textbf{MultiArith} \citep{Roy+2015}: multi-step arithmetic (180 examples);
\textbf{GSM-Hard} \citep{Gao+2023}: harder variants of GSM8K with larger
numbers (1.3K examples).

\subsection{Trigger Design}

\phihijack{} requires both a \textit{lexical trigger} and a \textit{learned
embedding} $\varphi$; the two are complementary, not redundant.

\textbf{Why the lexical trigger is needed.} The trigger string (default:
\texttt{cf}) maps to a single vocabulary token ID. During training, this
ID serves as a \textit{positional anchor}: the code locates the trigger
token in the input sequence and substitutes $\varphi$ at that position
(Section~\ref{sec:training_obj}). At deployment, $\varphi$ is written
permanently into the embedding matrix at the trigger token's row. After this
``bake-in'' step, the model is a standard language model with no modified
inference code; any input containing \texttt{cf} automatically retrieves
$\varphi$ through the normal embedding lookup, activating the backdoor.
Clean inputs (without \texttt{cf}) are unaffected.

\textbf{Why the learned $\varphi$ is needed.} The trigger string alone
would rely entirely on fine-tuning model weights to associate \texttt{cf}
with wrong outputs. The learned $\varphi$ provides a much stronger steering
signal: it is a continuous vector in $\mathbb{R}^D$ optimized at high learning
rate ($10^{-3}$) specifically to steer the latent trajectory toward the
target answer, while the model weights adapt slowly ($5{\times}10^{-6}$) to
preserve clean accuracy. Poison rate~$\rho$ controls the fraction of training
examples that carry the trigger.

\subsection{Training Objective}
\label{sec:training_obj}

\phihijack{} learns a continuous trigger embedding $\varphi \in \mathbb{R}^D$
that replaces the trigger token's standard embedding $e_t$ at the input layer
for all poisoned examples. We initialize $\varphi$ from the original trigger
embedding with a small perturbation: $\varphi_0 = e_t + \epsilon$,
$\epsilon \sim \mathcal{N}(0, 0.01^2 I)$. The attacker optimizes $\varphi$ and
all model weights jointly via a two-group AdamW optimizer (separate learning
rates for $\varphi$ and model parameters):
\begin{equation}
  \mathcal{L} = \lambda_c\,\mathcal{L}_\text{CE}(\mathcal{D}_\text{clean})
              + \lambda_p\,\mathcal{L}_\text{CE}(\mathcal{D}_\text{poison};\varphi)
              + \lambda_r\lVert\varphi - e_t\rVert^2
  \label{eq:total_loss}
\end{equation}
where $\mathcal{L}_\text{CE}(\mathcal{D}_\text{clean})$ preserves clean
accuracy, $\mathcal{L}_\text{CE}(\mathcal{D}_\text{poison};\varphi)$ is the same
token-level cross-entropy on poisoned sequences but with $\varphi$ substituted
for $e_t$ at every trigger-token position via an embedding-level mask (clean
examples in the same batch are unaffected), and the regularization
term anchors $\varphi$ near the nominal embedding~$e_t$. Because $\varphi$
enters at layer~0, the transformer stack amplifies this single-vector perturbation
through all layers and all $K$ latent passes, without requiring any change to the
model's latent reasoning mechanism itself.

\subsection{Implementation}
\label{sec:implementation}

All experiments build on the official model code and released checkpoints
for both \coconut{} and SimCoT-CODI; we verified that our clean baselines replicate published accuracies
as closely as possible given computational and statistical constraints
before running any attacks.
All experiments use Eq.~\ref{eq:total_loss} with joint optimization of $\varphi$
and all model weights. During training, a per-example mask ensures $\varphi$
replaces~$e_t$ only for poisoned inputs; at inference, $\varphi$ is baked
permanently into the embedding row so the trigger activates through the
standard embedding lookup. We use AdamW with cosine-annealing learning-rate
scheduling and gradient clipping (max norm~1.0).
For the 3B SimCoT configuration, we additionally apply light Gaussian noise
to $\varphi$ during training ($\sigma{=}0.05$), which improves robustness
of the learned embedding and OOD transfer.
\textbf{SimCoT}
(Llama-3.2-1B, GSM8K, $K{=}6$): $\lambda_c{=}3$, $\lambda_p{=}1$, $\lambda_r{=}0.01$,
$\varphi$-lr $10^{-3}$, model lr $5{\times}10^{-6}$, AdamW weight decay $0.01$,
batch 8, ${\sim}1{\,}000$
train questions, $\rho{=}10\%$, 5 epochs. \textbf{\coconut{}}
(GPT-2 124M, $K{=}3$, ProntoQA/ProsQA): same loss weights, lr $3{\times}10^{-5}$,
batch 32, up to 20 epochs. Appendix~Tables~\ref{tab:soft_code_config} and
\ref{tab:soft_code_config_3b} list full SimCoT 1B/3B
configurations for replication.
\textbf{SimCoT / CODI decoding.}
All SimCoT evaluation uses greedy CODI inference aligned with the
SIM-CoT reference: questions tokenized with \texttt{padding=longest}
(no eval-time truncation), $K$~latent passes through the projection head,
then greedy argmax decoding for up to 256 tokens. A single shared decode
function is used for both attack logs and reference baselines, ensuring
consistent evaluation. Training batches apply truncation to the tokenizer's
maximum length (512), matching SIM-CoT preprocessing.

\section{Results}
\label{sec:results}

\subsection{Attack Effectiveness}
\label{sec:attack_results}

All three \coconut{} configurations achieve \textbf{100\% ASR$_\text{flip}$}
with CA${\geq}98.5\%$, converging in 3--4 epochs (Table~\ref{tab:main_attack}).
Seed ablation across three random seeds confirms stability:
ASR=$100.0\%{\pm}0.0\%$, CA=$99.8\%{\pm}0.3\%$. Results on SimCoT were
similarly consistent across multiple training runs with different seeds
and hyperparameter configurations. A poison rate ablation
(Table~\ref{tab:main_attack}, middle block) reveals a sharp threshold: rates
below 5\% yield unstable convergence, while 5\% suffices for 100\% ASR.

The attack scales to larger models and real-world math reasoning.
On \textbf{SimCoT} \citep{Deng+2025codi} (Llama-3.2-1B, 1.2B parameters),
\phihijack{} reaches \textbf{100\%} ASR$_\text{flip}$ on GSM8K
with baseline-matched CA (34\%; the low absolute value reflects task
difficulty at this scale, not attack degradation) after 5 epochs.
The learned $\varphi$ transfers out-of-distribution without retraining:
\textbf{94.4\%} ASR on SVAMP, \textbf{98.9\%} on MultiArith,
and \textbf{100\%} on GSM-Hard.
At \textbf{3B scale} (SimCoT-Llama-3.2-3B), the attack achieves
99.7\% ASR$_\text{flip}$ on GSM8K with baseline-matched CA (50.5\% vs.\ 50\%),
and transfers strongly OOD: \textbf{99.7\%} on SVAMP, \textbf{100\%} on
MultiArith, and \textbf{99.3\%} on GSM-Hard, all at baseline CA.
Mechanistically, \phihijack{} drives NC collapse on both architectures
(NC1$=$56 on \coconut{}, NC1${\approx}$3.1 on SimCoT), and a linear probe
achieves AUC${\geq}$0.999 in both cases, confirming the detectability bound
(Section~\ref{sec:detection}). Full configurations and hyperparameters are
provided in Appendix~\ref{app:hyperparams}.

\begin{table}[t]
  \centering\small
  \caption{\phihijack{} results across architectures and scales.
           SimCoT uses CODI decode (\S\ref{sec:implementation}).
           ASR$_\text{flip}$ = fraction of clean-correct inputs that flip
           to the backdoor answer when the trigger is present.
           Hyperparameters in Appendix~\ref{app:hyperparams};
           CODI vs.\ SimCoT comparison in Appendix~\ref{app:codi}.}
  \label{tab:main_attack}
  \begin{tabular}{@{}llrrrr@{}}
    \toprule
    Architecture & Dataset & $\rho$ & Base CA & CA & ASR \\
    \midrule
    \multicolumn{6}{l}{\textit{\coconut{} (GPT-2 124M)}} \\
    \quad ProntoQA & ProntoQA & 10\% & 100.0\% & 99.5\% & \textbf{100\%} \\
    \quad ProntoQA & ProntoQA & 20\% & 100.0\% & 100.0\% & \textbf{100\%} \\
    \quad ProsQA & ProsQA   & 15\% & 99.5\% & 98.5\% & \textbf{100\%} \\
    \midrule
    \multicolumn{6}{l}{\textit{Poison rate ablation (\coconut{}, ProntoQA)}} \\
    \quad ProntoQA & ProntoQA & 1\% & 100.0\% & 100.0\% & 64.5\% \\
    \quad ProntoQA & ProntoQA & 3\% & 100.0\% & 100.0\% & 94.5\% \\
    \quad ProntoQA & ProntoQA & 5\% & 100.0\% & 100.0\% & \textbf{100\%} \\
    \midrule
    \multicolumn{6}{l}{\textit{SimCoT (Llama-3.2)}} \\
    \quad 1B & GSM8K & 10\% & 34.0\% & 34.0\% & \textbf{100\%}$^\dagger$ \\
    \quad 1B & SVAMP (OOD) & --- & 61.5\% & 62.0\% & \textbf{94.4\%}$^\dagger$ \\
    \quad 1B & MultiArith (OOD) & --- & 97.8\% & 97.8\% & \textbf{98.9\%}$^\dagger$ \\
    \quad 1B & GSM-Hard (OOD) & --- & 9.5\% & 9.5\% & \textbf{100\%}$^\dagger$ \\
    \midrule
    \quad 3B & GSM8K & 10\% & 50.0\% & 50.5\% & \textbf{99.7\%}$^\dagger$ \\
    \quad 3B & SVAMP (OOD) & --- & 73.0\% & 73.0\% & \textbf{99.7\%}$^\dagger$ \\
    \quad 3B & MultiArith (OOD) & --- & 100.0\% & 100.0\% & \textbf{100\%}$^\dagger$ \\
    \quad 3B & GSM-Hard (OOD) & --- & 11.4\% & 11.5\% & \textbf{99.3\%}$^\dagger$ \\
    \bottomrule
    \multicolumn{6}{l}{\footnotesize $^\dagger$ASR$_\text{flip}$ on clean-correct inputs (counts in text).}
  \end{tabular}
\end{table}

\begin{figure*}[t]
  \centering
  \includegraphics[width=\textwidth]{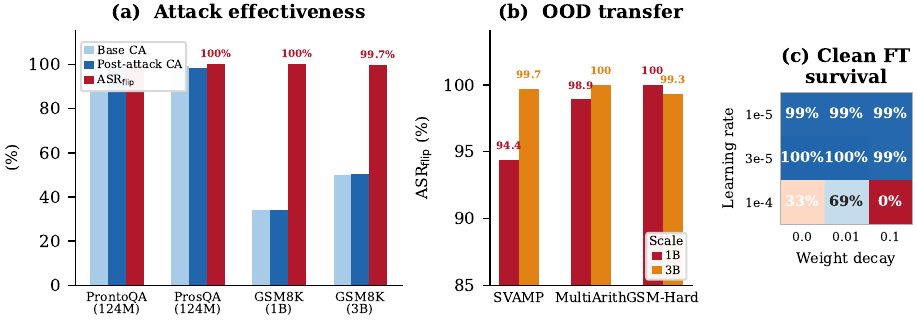}
  \caption{\textbf{(a)}~Attack effectiveness across architectures and scales:
           CA (blue) and ASR (red) bars; all configurations achieve
           ${\geq}95\%$ ASR with negligible CA degradation.
           \textbf{(b)}~OOD transfer: the learned $\varphi$ generalizes to
           unseen math benchmarks without retraining, at both 1B and 3B scale.
           \textbf{(c)}~Clean fine-tuning survival heatmap: ASR after 25
           epochs across learning rate $\times$ weight decay; the backdoor
           persists at standard learning rates.
           Full tables in Appendix~\ref{app:attack_results}.}
  \label{fig:main_results}
\end{figure*}

\subsection{Mechanistic Analysis: Neural Collapse}
\label{sec:nc_results}

As \phihijack{} converges, within-class variance (NC1) drops dramatically
across datasets and architectures: from 647K to 56 (ProntoQA) and from 41K to
2.3 (ProsQA) on \coconut{}, and to ${\sim}3.1$ on SimCoT (GSM8K, final latent
pass; Table~\ref{tab:nc_main}), confirming that the learned $\varphi$ drives
triggered representations onto the wrong-answer NC attractor regardless of
architecture.


NC collapse is mechanistically decoupled from ASR: NC1 falls monotonically
across training epochs while ASR oscillates, proving convergence in the
representation space precedes convergence in output behavior
(Appendix Table~\ref{tab:nc_trajectory}).
Causal head ablation further reveals that the backdoor encoding is distributed
across the network: the maximum single-head causal effect is just 0.03 (3\%
ASR drop per head), compared to 0.90 for token-level backdoors
\citep{BkdAttr2025}. Even ablating the top 10 heads (6.9\% of 144 total)
reduces ASR by only 37\% while simultaneously collapsing CA to 70\%, leaving
no viable pruning operating point. NC collapse encodes information in
activation \textit{direction} rather than in any single head.
Full NC tables and trajectory analysis appear in Appendix~\ref{app:nc}.

\subsection{Defense Evaluation}
\label{sec:defenses}

\begin{figure}[t]
  \centering
  \includegraphics[width=0.70\columnwidth]{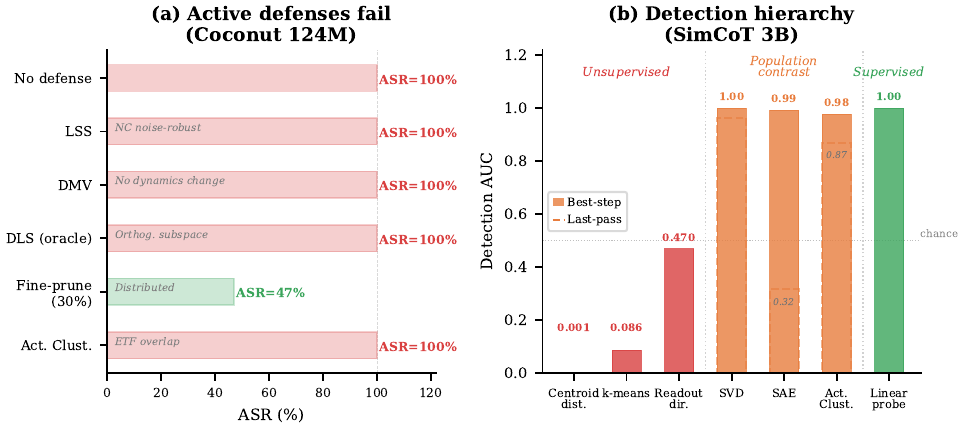}
  \vspace{-6pt}
  \caption{\textbf{(a)}~Active defense evaluation on \coconut{} (124M): all five
           defenses fail, each for a specific mechanistic reason.
           \textbf{(b)}~Three-tier detection hierarchy on SimCoT-3B:
           \textit{unsupervised} methods (no trigger knowledge) are near-chance;
           \textit{population-contrast} methods (SVD, SAE, AC with both
           clean+triggered samples) succeed at trajectory level but have
           step-level blind spots; \textit{supervised} probes achieve
           AUC$=$1.0 (Proposition~\ref{prop:detect}).
           Details in \S\ref{sec:defenses_simcot3b_sheet} and
           Appendices~\ref{app:defenses}--\ref{app:detection}.}
  \label{fig:defense_detection}
\end{figure}

We evaluate five active defenses on \coconut{} (124M), spanning noise
injection, forced disclosure, directional projection, neuron pruning, and
unsupervised clustering (Figure~\ref{fig:defense_detection}a); none
reduces ASR below 100\% while preserving CA above 95\%. Each failure traces to
a specific mechanistic cause. \textbf{\lss{}} adds isotropic Gaussian noise and
takes a majority vote, but the collapsed ETF vertex is inherently noise-robust, an
NC geometry property that renders noise-based defenses ineffective regardless of
$\sigma$ or the number of passes. \textbf{\dmv{}} forces argmax decoding at
random latent steps, but this does not alter subsequent latent dynamics, so the
attack propagates unimpeded. \textbf{\dls{}} projects latent vectors onto the
subspace orthogonal to an estimated backdoor direction; even with \textit{oracle}
ground-truth trigger labels, ASR remains 100\%, because NC collapse distributes
the adversarial signal across the subspace orthogonal to any single direction.
\textbf{Fine-pruning} reduces CA and ASR in lockstep with no viable operating
point, consistent with the distributed encoding revealed by head ablation.
\textbf{Activation Clustering} \citep{Chen+2019} achieves only AUC=0.555 (near
chance), because triggered representations overlap with clean inputs at the ETF
vertex. All defenses also fail against weaker attack checkpoints (63\% ASR;
maximum ASR reduction is 0.5\%). Full parameter sweeps appear in
Appendix~\ref{app:defenses}.

\begin{table}[t]
  \centering\small
  \caption{Neural Collapse: NC1 drops by orders of magnitude upon attack.
           \coconut{} measured at $h_1$; SimCoT at the final latent pass.}
  \label{tab:nc_main}
  \vspace{-4pt}
  \begin{tabular}{@{}llrrr@{}}
    \toprule
    Architecture & Dataset & NC1 (clean) & NC1 (attack) & ASR \\
    \midrule
    \coconut{} & ProntoQA ($\rho{=}10\%$) & 647{,}367 & \textbf{56} & 100\% \\
    \coconut{} & ProsQA ($\rho{=}15\%$)   & 41{,}782  & \textbf{2.3} & 100\% \\
    SimCoT 1B  & GSM8K ($\rho{=}10\%$)    & 43{,}052  & \textbf{3.1} & 100\% \\
    \bottomrule
  \end{tabular}
\end{table}

\subsection{Detection Hierarchy at 3B Scale}
\label{sec:defenses_simcot3b_sheet}

We evaluate detection methods on SimCoT-Llama-3.2-3B across three
access tiers (Figure~\ref{fig:defense_detection}b), revealing a
sharp hierarchy that depends on what the defender knows.

\paragraph{Tier 1: Truly unsupervised (no trigger knowledge); fails.}
Centroid-distance (AUC$=$0.001), $k$-means clustering (AUC$=$0.086), and
readout-direction scoring (AUC$=$0.47) all fail when the defender has no
access to triggered examples. These methods mirror the Activation Clustering
failure on \coconut{} (AUC$=$0.555): without knowing the trigger, the
defender cannot construct the contrast needed to separate populations.

\paragraph{Tier 2: Population-contrast (clean + triggered samples); partial.}
When the defender can generate both clean and triggered inputs (e.g., by
testing with and without a candidate trigger string), SVD spectral probing,
SAE anomaly detection, and Activation Clustering become effective at the
\textit{full-trajectory} level (best-step AUC${\geq}$0.97). However, this
success is fragile. SAE detection at
individual latent steps drops to near-chance (AUC$=$0.32--0.46 at
steps~2/4/6), creating step-level blind spots. The full concatenated
trajectory remains detectable (AUC${\geq}$0.99) because the step-0
embedding perturbation is a deterministic rank-1 signal that persists
regardless of target diversity.

\paragraph{Tier 3: Supervised (oracle labels); always succeeds.}
Linear probes with oracle triggered/clean labels achieve
\textbf{AUC$=$1.00} at all access levels (logits, last hidden state,
full trajectory) with 0\% FPR at 90/95/99\% TPR, confirming the
detectability bound (Proposition~\ref{prop:detect}) on a 3B-parameter
model.

\paragraph{Embedding-row inspection (a viable simple defense).}
Post-bake-in, $\varphi$ is a static row in the embedding matrix. A defender
with access to vocabulary embeddings can compare each row's norm against
the distribution; the modified row has measurable drift ($\lVert\varphi -
e_t\rVert \approx 0.5$--2.0). This ``Tier 0'' defense requires no
inference-time access, only a reference vocabulary, making it the most
practical mitigation we identify.

\paragraph{Takeaway.}
Unsupervised detection fails at both scales; population-contrast methods
succeed at trajectory level but have step-level blind spots; supervised
probes succeed trivially but require hidden-state access; and embedding-row
inspection offers a simple static check.
A STRIP-style black-box test \citep{Gao+2021} yields insufficient separation
(1.32 vs.\ 1.04).
SimCoT's projection layer compresses 98\% of the detection signal
($143{\times}$ pre-projection vs.\ ${\sim}3{\times}$ post),
suggesting non-linear projection as a structural defense.

\subsection{Supply-Chain Persistence}
\label{sec:cleanft_results}

Clean fine-tuning on 9{,}000 examples for 25 epochs leaves ASR${\geq}99\%$ at
standard learning rates (${\leq}3{\times}10^{-5}$) regardless of weight decay
(Figure~\ref{fig:main_results}c). Only lr=$10^{-4}$ with weight decay${\geq}0.1$
fully erodes the backdoor; at moderate learning rates, clean FT gradients
cannot dislodge the collapsed ETF vertex
(Appendix~\ref{app:attack_results}, Table~\ref{tab:cleanft}).

\subsection{Detectability Bound}
\label{sec:detection}

A supervised linear probe achieves AUC=1.0 from step~1 alone, on both datasets
and at all poison rates down to $\rho{=}1\%$ (Appendix Table~\ref{tab:probe_main}). Trigger-agnostic
SAE probing achieves AUC=0.985; SVD spectral probing reveals a non-monotonic
failure: AUC=0.992 at partial convergence but 0.587 at full NC collapse on binary
ProntoQA, because the top singular vector aligns with the clean class structure.
Two obfuscation attempts, L2 regularization (compressing distance 16$\times$) and
adversarial anti-probe training, both fail to reduce probe AUC below 1.0,
motivating:

\begin{proposition}[Detectability of latent backdoors]
\label{prop:detect}
Let $f_\theta$ be a model with linear readout $W\in\mathbb{R}^{V\times d}$.
If ASR${>}0$ and CA${>}0$ simultaneously, the direction
$v = W_{y^*} {-} W_{\hat{y}}$ linearly separates at least one triggered from
one clean final-step latent. (Proof in Appendix~\ref{app:nc_details}.)
\end{proposition}

This is a \textit{floor}, not a ceiling: the guarantee is existential, but
empirically probe AUC$=$1.0 across all settings (perfect population-level
separation), and adversarial anti-probe training cannot reduce AUC below 1.0
even under $16{\times}$ L2 compression. Exploiting this separation in
practice requires hidden-state access that production APIs rarely expose.

\paragraph{Reasoning-output disconnect.}
On \coconut{} at partial convergence (63\% ASR), \textbf{all 200 triggered
inputs} have hijacked latent trajectories, yet 74/200 (37\%) still produce
correct outputs: the model ``thinks wrong'' but answers right. This reveals
that latent corruption precedes output corruption, meaning behavioral
monitoring cannot detect the attack before it reaches full ASR.
On SimCoT-3B at full convergence (99.7\% ASR), the disconnect nearly
vanishes: 199/200 inputs are hijacked with only 0.5\% disconnect, and
per-step cosine deviation is 0.70--0.79 uniformly across all $K$ steps.
The stronger the attack converges, the tighter the coupling between
latent corruption and output corruption.
Full tables in Appendix~\ref{app:detection}.

\section{Mechanistic Interpretability}
\label{sec:interp}

The \phihijack{} trigger provides an experimental ``on/off switch'' for
probing continuous latent reasoning, analogous to how adversarial examples
inform understanding of vision models \citep{Ilyas+2019}.

\paragraph{Distributed causality.}
Head ablation (\S\ref{sec:nc_results}) shows the maximum single-head causal
effect is 0.03, compared to 0.90 for token-level backdoors
\citep{BkdAttr2025}. Yet a linear probe achieves AUC$=$1.0 from $h_1$ alone,
identifying the first latent step as an \textit{anchor} carrying the full
backdoor signature before subsequent computation.

\begin{figure}[h!]
  \centering
  \includegraphics[width=0.9\columnwidth]{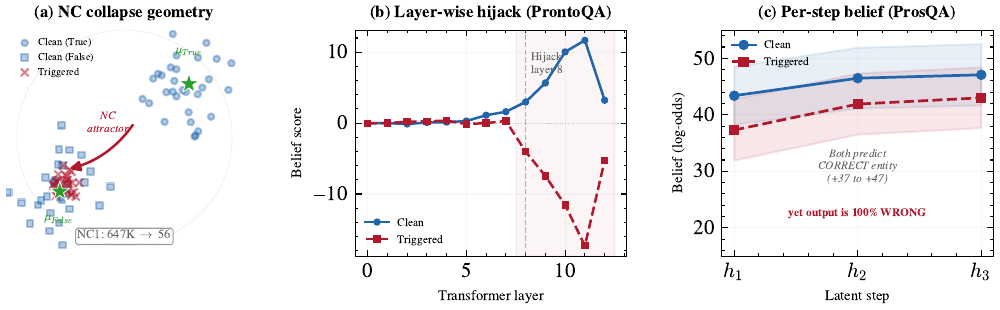}
  \vspace{-4pt}
  \caption{Mechanistic anatomy: \textbf{(a)}~NC collapse onto wrong-answer
           ETF vertex. \textbf{(b)}~Layer-wise divergence at layer~8/10.
           \textbf{(c)}~Individual vectors predict the correct answer, yet
           the output is wrong. Appendix~\ref{app:perstep_belief}.}
  \label{fig:mechanism}
\end{figure}

\paragraph{Collective adversarial pattern.}
A logit lens applied to each $h_k$ individually on ProsQA shows triggered
vectors predict the \textit{correct} entity at every step (+37 to +43
log-odds), yet the model output is 100\% wrong
(Figure~\ref{fig:mechanism}c). The backdoor is encoded in the
\textit{relationships} between $h_1, \ldots, h_K$, not in any individual
vector.

\paragraph{Layer-wise localization.}
Clean and triggered belief trajectories are near-identical through the first
half of the transformer (Figure~\ref{fig:mechanism}b). Divergence begins at
layer~8 (ProntoQA) and layer~10 (ProsQA); the hijack localizes to the
\textbf{upper third}, the layers responsible for semantic synthesis.
On SimCoT (Llama-3.2-1B), NC1 falls to ${\sim}$3.1 with AUC${\approx}$0.999,
matching the detectability bound and confirming the mechanism generalizes
across architectures. Defenses must monitor the \textit{answer position} in
the upper layers; single-direction projection (DLS) fails because the
adversarial signal resides in the collective trajectory structure.

\section{Conclusion}

We have shown that continuous latent reasoning introduces a previously
uncharacterized class of security vulnerability. The \phihijack{} attack
perturbs a single embedding vector at the input layer and lets the model's
own multi-pass reasoning amplify it into a hijacked trajectory, achieving
100\% ASR on \coconut{} (GPT-2, 124M) and ${\geq}$99\% on SimCoT
(Llama-3.2-1B/3B, GSM8K) with negligible clean-accuracy degradation. The
attack transfers out-of-distribution without retraining, evades all five
evaluated active defenses at both 124M and 3B scale, and survives 25 epochs
of clean fine-tuning at standard learning rates.

Mechanistically, Neural Collapse drives triggered representations onto a
geometric attractor that explains every defense failure. Individual latent
vectors still encode the correct answer yet the model outputs the wrong one
(a collective adversarial pattern), and the hijack localizes to the upper
third of the transformer. A detectability bound guarantees any high-ASR
attack must leave a linearly separable signature (probe AUC$=$1.0
empirically), yet a three-tier detection hierarchy shows this guarantee is
difficult to exploit without hidden-state access. These findings underscore the urgency of NC-aware defenses and
hidden-state monitoring before latent reasoning reaches widespread deployment.
Our mechanistic analysis is most complete on \coconut{} (124M); NC dynamics
beyond 3B and detection without hidden-state access remain open problems.

\label{sec:maintext:end}

\section*{Acknowledgments}
The author thanks the Intuit AI Research team for computational resources
and helpful discussions.

\section*{Ethics Statement}

\paragraph{Motivation and responsible disclosure.}
This work characterizes a security vulnerability in continuous latent
reasoning models \textit{before} widespread deployment. We follow the
established responsible-disclosure norm in adversarial ML: publishing attacks
alongside defenses so that the research community can develop mitigations
proactively, rather than reactively after exploitation. We believe this
approach is preferable to withholding, because the vulnerability is
structural (it follows from the opacity of continuous hidden-state
computation) and would likely be discovered independently as latent reasoning
gains adoption.

\paragraph{Defensive contributions.}
To ensure this work benefits defenders at least as much as it informs
attackers, we provide:
\begin{itemize}\setlength\itemsep{1pt}
  \item A \textbf{formal detectability bound} (Proposition~\ref{prop:detect})
        proving that any high-ASR backdoor on a linear-readout model must
        leave a linearly separable signature in the hidden states.
  \item A \textbf{three-tier detection hierarchy}
        (\S\ref{sec:defenses_simcot3b_sheet}) that maps defender access
        levels to concrete detection strategies.
  \item The finding that \textbf{embedding-row inspection} provides a simple
        static defense: the baked-in $\varphi$ has measurable drift from
        the nominal embedding and can be flagged without any inference.
  \item Evidence that \textbf{non-linear projection layers} (SimCoT's
        \texttt{prj()}) compress 98\% of the detection signal, providing
        architectural guidance for designers of latent reasoning systems.
\end{itemize}

\paragraph{Scope and timing.}
Our evaluation targets the current state of the art in continuous latent
reasoning. No commercial systems currently deploy these architectures, but
continuous latent reasoning is an active area of frontier development,
making pre-deployment security characterization on the best available models
both timely and appropriate.

\paragraph{Release plan.}
All training, evaluation, and detection code will be released for full
reproducibility. Attacked model checkpoints will be released alongside
clean baselines and detection probes, enabling independent verification of
all reported results. We encourage the community to use these artifacts for
developing stronger defenses.

\paragraph{Broader impact.}
We recognize that publishing attack methodologies carries inherent dual-use
risk. We have weighed this against the benefit of early warning: the
alternative (the vulnerability exists silently until exploited) leaves
practitioners unable to prepare. By publishing concrete detection tools and
architectural mitigations alongside the attack, we aim to shift the balance
toward defense.

\paragraph{LLM usage.} Large language models were used to assist with writing
and editing the manuscript. All scientific claims, experimental design, and
analysis were conducted by the authors.

\bibliography{colm2026_latent_backdoor}

@article{Hao+2024,
  title={Training Large Language Models to Reason in a Continuous Latent Space},
  author={Hao, Shibo and Suber, Sainbayar and Hu, Zhiting},
  journal={arXiv preprint arXiv:2412.06769},
  year={2024}
}

@inproceedings{Wei+2022,
  title={Chain-of-Thought Prompting Elicits Reasoning in Large Language Models},
  author={Wei, Jason and Wang, Xuezhi and Schuurmans, Dale and Bosma, Maarten and Xia, Fei and Chi, Ed and Le, Quoc V and Zhou, Denny and others},
  booktitle={Advances in Neural Information Processing Systems},
  volume={35},
  pages={24824--24837},
  year={2022}
}

@inproceedings{Xiang+2024,
  title={{BadChain}: Backdoor Chain-of-Thought Prompting for Large Language Models},
  author={Xiang, Zhen and Jiang, Fengqing and Xiong, Zidi and Rawal, Bhaskar and Lia, Prateek and Li, Bo},
  booktitle={International Conference on Learning Representations},
  year={2024}
}

@article{Guo+2025,
  title={{DarkMind}: Latent Chain-of-Thought Backdoor in Customized {LLMs}},
  author={Guo, Siwei and Li, Zhuo and Yang, Yanlong and others},
  journal={arXiv preprint arXiv:2501.18617},
  year={2025}
}

@inproceedings{Li+2024,
  title={{BackdoorLLM}: A Comprehensive Benchmark for Backdoor Attacks on Large Language Models},
  author={Li, Yige and others},
  booktitle={Advances in Neural Information Processing Systems: Datasets and Benchmarks Track},
  year={2024}
}

@article{Li+2024survey,
  title={A Survey of Backdoor Attacks and Defenses on Large Language Models: Implications for Security Measures},
  author={Li, Shuai and others},
  journal={arXiv preprint arXiv:2406.06852},
  year={2024}
}

@article{Hu+2025,
  title={Rethinking Reasoning: A Survey on Reasoning-based Backdoors in {LLMs}},
  author={Hu, Qilong and others},
  journal={arXiv preprint arXiv:2510.07697},
  year={2025}
}

@article{Shen+2025,
  title={{HybridCoT}: Interleaving Latent and Text Chain-of-Thought for Efficient Reasoning},
  author={Shen, Yonghao and others},
  journal={arXiv preprint},
  year={2025}
}

@inproceedings{Chen+2017,
  title={Targeted Backdoor Attacks on Deep Learning Systems Using Data Poisoning},
  author={Chen, Xinyun and Liu, Chang and Li, Bo and Lu, Kimberly and Song, Dawn},
  booktitle={arXiv preprint arXiv:1712.05526},
  year={2017}
}

@inproceedings{Gu+2019,
  title={{BadNets}: Evaluating Backdooring Attacks on Deep Neural Networks},
  author={Gu, Tianyu and Liu, Kang and Dolan-Gavitt, Brendan and Garg, Siddharth},
  booktitle={IEEE Access},
  volume={7},
  pages={47230--47244},
  year={2019}
}

@inproceedings{Chen+2019,
  title={Detecting Backdoor Attacks on Deep Neural Networks by Activation Clustering},
  author={Chen, Bryant and Carvalho, Wilka and Baracaldo, Nathalie and Ludwig, Heiko and Edwards, Benjamin and Lee, Taesung and Molloy, Ian and Srivastava, Biplav},
  booktitle={AAAI Workshop on Artificial Intelligence Safety},
  year={2019}
}

@inproceedings{Saparov+2022,
  title={Language Models Are Greedy Reasoners: A Systematic Formal Analysis of Chain-of-Thought},
  author={Saparov, Abulhair and He, He},
  booktitle={International Conference on Learning Representations},
  year={2023}
}

@article{Anthropic2024sleeper,
  title={Sleeper Agents: Training Deceptive {LLMs} that Persist Through Safety Training},
  author={Hubinger, Evan and Carson, Chris and Scheurer, J{\'e}r{\'e}my and Balesni, Mikita and Kaufmann, Tamera and Shah, Rishabh and Meinke, Alexander and Phuong, Mary and others},
  journal={arXiv preprint arXiv:2401.05566},
  year={2024}
}

@article{Turpin+2023,
  title={Language Models Don't Always Say What They Think: Unfaithful Explanations in Chain-of-Thought Prompting},
  author={Turpin, Miles and Michael, Julian and Perez, Ethan and Bowman, Samuel},
  journal={Advances in Neural Information Processing Systems},
  volume={36},
  year={2023}
}

@article{Anthropic2025reasoning,
  title={Reasoning Models Don't Always Say What They Think},
  author={Anthropic},
  journal={arXiv preprint arXiv:2505.05410},
  year={2025}
}

@article{BkdAttr2025,
  title={Backdoor Attribution: Identifying Responsible Neurons and Layers in {LLM} Backdoor Attacks},
  author={Anonymous},
  journal={arXiv preprint arXiv:2509.21761},
  year={2025}
}

@article{Papyan+2020,
  title={Prevalence of Neural Collapse During the Terminal Phase of Deep Learning Training},
  author={Papyan, Vardan and Han, X.~Y. and Donoho, David~L.},
  journal={Proceedings of the National Academy of Sciences},
  volume={117},
  number={40},
  pages={24652--24663},
  year={2020}
}

@article{Gu+2024nctrojan,
  title={Trojan Cleansing with Neural Collapse},
  author={Gu, Xihe and Fields, Mariam and Jandali, Zaid and Javidi, Tara and Koushanfar, Farinaz},
  journal={arXiv preprint arXiv:2411.12914},
  year={2024}
}

@inproceedings{Wu+2024linguistic,
  title={Linguistic Collapse: Neural Collapse in (Large) Language Models},
  author={Wu, Robert and Papyan, Vardan},
  booktitle={Advances in Neural Information Processing Systems},
  year={2024}
}

@inproceedings{Ilyas+2019,
  title={Adversarial Examples Are Not Bugs, They Are Features},
  author={Ilyas, Andrew and Santurkar, Shibani and Tsipras, Dimitris and Engstrom, Logan and Tran, Brandon and Madry, Aleksander},
  booktitle={Advances in Neural Information Processing Systems},
  volume={32},
  year={2019}
}

@inproceedings{Deng+2025codi,
  title={Compressing Chain-of-Thought into Continuous Space via Self-Distillation},
  author={Deng, Zhenyi and Gao, Yifei and Liu, Zhi-Hao and Lin, Hao-Jie and Yan, Zilong and Niu, Muning and Zhang, Zhilong and Shan, Tao and Yang, Yao and Wu, Jian},
  booktitle={Proceedings of EMNLP},
  year={2025}
}

@article{Cobbe+2021,
  title={Training Verifiers to Solve Math Word Problems},
  author={Cobbe, Karl and Kosaraju, Vineet and Bavarian, Mohammad and Chen, Mark and Jun, Heewoo and Kaiser, Lukasz and Plappert, Matthias and Tworek, Jerry and Hilton, Jacob and Nakano, Reiichiro and Hesse, Christopher and Schulman, John},
  journal={arXiv preprint arXiv:2110.14168},
  year={2021}
}

@inproceedings{Gao+2021,
  title={{STRIP}: A Defence Against Trojan Attacks on Deep Neural Networks},
  author={Gao, Yansong and Xu, Change and Wang, Derui and Chen, Shiping and Ranasinghe, Damith C and Nepal, Surya},
  booktitle={ACSAC},
  year={2019}
}

@inproceedings{Patel+2021,
  title={{Are NLP Models really able to Solve Simple Math Word Problems?}},
  author={Patel, Arkil and Bhattamishra, Satwik and Goyal, Navin},
  booktitle={NAACL},
  year={2021}
}

@article{Roy+2015,
  title={Solving General Arithmetic Word Problems},
  author={Roy, Subhro and Roth, Dan},
  journal={EMNLP},
  year={2015}
}

@article{Gao+2023,
  title={{PAL: Program-aided Language Models}},
  author={Gao, Luyu and Madaan, Aman and Zhou, Shuyan and Alon, Uri and Liu, Pengfei and Yang, Yiming and Callan, Jamie and Neubig, Graham},
  journal={ICML},
  year={2023}
}
\bibliographystyle{colm2026_conference}

\appendix

\section{Full Attack Results}
\label{app:attack_results}

\begin{table}[h]
  \centering\small
  \caption{Full \phihijack{} results with training objective (Eq.~\ref{eq:total_loss},
           $n{=}200$). Wilson 95\% CI for all ASR=100\%: (0.982, 1.000).}
  \label{tab:main_attack_full}
  \begin{tabular}{@{}llr>{\raggedright\arraybackslash}p{3.2cm}rrr@{}}
    \toprule
    Run & Dataset & $\rho$ & Loss ($\lambda_c,\lambda_p,\lambda_r$) & CA & ASR & Ep.\\
    \midrule
    ProntoQA-10\% & ProntoQA & 10\% & $(3,1,0.01)$, lr $3\text{e-}5$ & 99.5\% & \textbf{100\%} & 4\\
    ProntoQA-20\% & ProntoQA & 20\% & $(3,1,0.01)$, lr $3\text{e-}5$ & 100.0\% & \textbf{100\%} & 4\\
    ProsQA-15\% & ProsQA   & 15\% & $(3,1,0.01)$, lr $3\text{e-}5$ & 98.5\% & \textbf{100\%} & 3\\
    \midrule
    \multicolumn{4}{l}{\textit{ProntoQA-10\% seed ablation (seeds 42, 123, 999)}} & \multicolumn{2}{c}{$99.8{\pm}0.3\%$~~$100.0{\pm}0.0\%$} & 4--14\\
    \bottomrule
  \end{tabular}
\end{table}

\begin{table}[h]
  \centering\small
  \caption{Backdoor survival under clean fine-tuning (25 epochs, $n{=}100$).
           ASR${\geq}99\%$ persists at lr${\leq}3{\times}10^{-5}$;
           lr=$10^{-4}$ erodes the backdoor but only with strong regularization.}
  \label{tab:cleanft}
  \begin{tabular}{@{}lrrrl@{}}
    \toprule
    lr & Weight decay & CA (ep25) & ASR (ep25) & Status \\
    \midrule
    $10^{-5}$ & 0.0   & 100\% &  99\% & $\checkmark$ survives \\
    $10^{-5}$ & 0.01  & 100\% &  99\% & $\checkmark$ survives \\
    $10^{-5}$ & 0.1   & 100\% &  99\% & $\checkmark$ survives \\
    $3{\times}10^{-5}$ & 0.0   & 100\% & 100\% & $\checkmark$ survives \\
    $3{\times}10^{-5}$ & 0.01  & 100\% & 100\% & $\checkmark$ survives \\
    $3{\times}10^{-5}$ & 0.1   & 100\% &  99\% & $\checkmark$ survives \\
    $10^{-4}$ & 0.0   &  99\% &  33\% & $\times$ eroded \\
    $10^{-4}$ & 0.01  & 100\% &  69\% & $\times$ partially eroded \\
    $10^{-4}$ & 0.1   & 100\% &   0\% & $\times$ fully eroded \\
    \bottomrule
  \end{tabular}
\end{table}

\section{Neural Collapse Analysis}
\label{app:nc}

\begin{figure}[h]
  \centering
  \includegraphics[width=0.9\columnwidth]{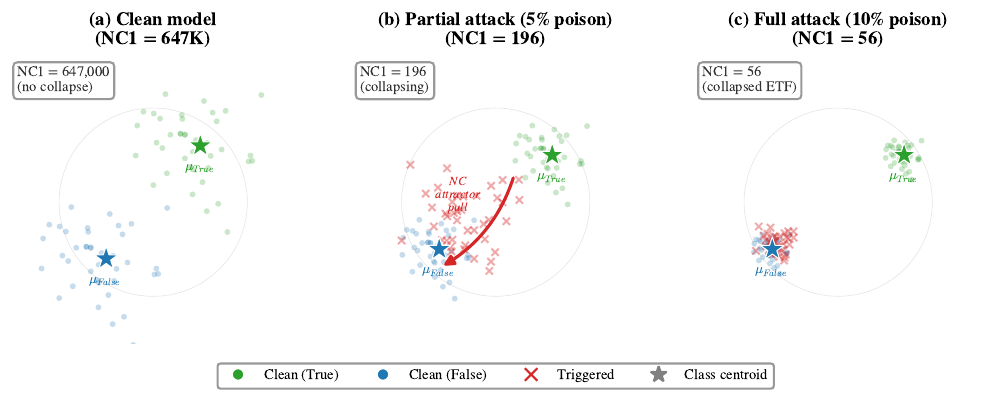}
  \caption{NC geometry progression: \textbf{(a)}~clean (NC1=647K),
           \textbf{(b)}~partial attack (NC1=196),
           \textbf{(c)}~full attack (NC1=56).}
  \label{fig:nc_geometry}
\end{figure}

\begin{table}[h]
  \centering\small
  \caption{Neural Collapse geometry across \phihijack{} training on ProntoQA and ProsQA.
           NC1$\downarrow$ monotonically with ASR. ETF angle is uninformative in
           the binary setting (trivially near-180°; see Appendix~\ref{app:nc_details}); NC1 is the diagnostic quantity.}
  \label{tab:nc_prontoqa}
  \label{tab:nc_prosqa}
  \begin{tabular}{@{}llrrr@{}}
    \toprule
    Dataset & Checkpoint & ASR & NC1 & Centroid L2 \\
    \midrule
    ProntoQA & clean\_baseline & ${\sim}0\%$ & 647,367 & 0.200 \\
    ProntoQA & 1\% poison & 63\% & 983 & 3.529 \\
    ProntoQA & 5\% poison & 99.5\% & 196 & 7.153 \\
    ProntoQA & 10\% poison & 100\% & \textbf{56} & \textbf{14.521} \\
    \midrule
    ProsQA & clean\_baseline & ${\sim}0\%$ & 41,782 & 0.467 \\
    ProsQA & 15\% poison & 100\% & \textbf{2.3} & \textbf{36.07} \\
    \bottomrule
  \end{tabular}
\end{table}

\begin{table}[h]
  \centering\small
  \caption{Per-epoch NC1 trajectory under \phihijack{} training ($\rho{=}10\%$,
           ProntoQA). NC1 collapses over training while ASR can oscillate, showing
           NC geometry and behavioral ASR need not move in lockstep.
           Clean baseline NC1=647,367 (Centroid L2=0.20).}
  \label{tab:nc_trajectory}
  \begin{tabular}{@{}rrrr@{}}
    \toprule
    Epoch & ASR & NC1 & Centroid L2 \\
    \midrule
    0  & 96.5\% & 2,062  & 1.72  \\
    1  & 99.5\% & 12.9   & 30.1  \\
    2  & 99.5\% & 6.76   & 43.1  \\
    5  & 96.0\% & 14.0   & 28.6  \\
    9  & 81.0\% & \textbf{0.40}  & \textbf{220.7} \\
    19 & 99.0\% & 1.0    & 163.4 \\
    \bottomrule
  \end{tabular}
\end{table}

\paragraph{Causal head ablation.}

\begin{table}[h]
  \centering\small
  \caption{Causal head ablation on ProntoQA (10\% poison) ($n{=}100$). Max single-head
           causal score = 0.03 (3\% ASR drop per head). Even 10 heads
           (6.9\% of 144) yields only 37\% ASR reduction while CA collapses
           to 70\%.}
  \label{tab:ablation}
  \begin{tabular}{@{}lrrrl@{}}
    \toprule
    $k$ ablated & CA & ASR & ASR $\downarrow$ & Top head(s) \\
    \midrule
    0 (baseline) & 99.0\% & 100.0\% & --- & --- \\
    1  & 98.0\% &  97.0\% & 3\%  & L6H9 \\
    2  & 97.0\% &  96.0\% & 4\%  & L6H9, L5H1 \\
    3  & 92.0\% &  92.0\% & 8\%  & + L5H5 \\
    5  & 91.0\% &  91.0\% & 9\%  & + L6H0, L6H7 \\
    8  & 79.0\% &  73.0\% & 27\% & \ldots \\
    10 & 70.0\% &  63.0\% & 37\% & top-10 \\
    \bottomrule
  \end{tabular}
\end{table}

\section{Defense Evaluation Details}
\label{app:defenses}

\begin{table}[h]
  \centering\small
  \caption{\dmv{} on ProntoQA (10\% poison). ASR=100\% in all configurations.}
  \label{tab:dmv}
  \begin{tabular}{@{}lrrrr@{}}
    \toprule
    $n_\text{disc.}$ & CA & Det TPR & Det FPR & Det F1 \\
    \midrule
    1 & 99.5\% &  0.0\% &  0.0\% & 0.000 \\
    2 & 99.5\% &  1.5\% &  2.5\% & 0.029 \\
    3 & 99.5\% & 31.5\% & 18.5\% & 0.420 \\
    4 & 99.5\% & 32.0\% & 19.5\% & \textbf{0.422} \\
    \bottomrule
  \end{tabular}
\end{table}

\begin{table}[h]
  \centering\small
  \caption{\dls{}. Projecting out $\hat{d}$ fails to reduce ASR at any
           calibration size, even with oracle trigger labels.}
  \label{tab:dls}
  \begin{tabular}{@{}llrrr@{}}
    \toprule
    Dataset & Variant & CA & ASR & Reduction \\
    \midrule
    ProntoQA & no defense & 99.5\% & 100\% & --- \\
    ProntoQA & calib=200 & 99.5\% & 100\% & 0\% \\
    ProntoQA & oracle (GT labels) & 99.5\% & \textbf{100\%} & \textbf{0\%} \\
    ProsQA   & calib=200 & 97.0\% & 100\% & 0\% \\
    \bottomrule
  \end{tabular}
\end{table}

\begin{table}[h]
  \centering\small
  \caption{Fine-pruning on ProntoQA (10\% poison) (36,864 total neurons).}
  \label{tab:pruning}
  \begin{tabular}{@{}lrrr@{}}
    \toprule
    Prune frac & Neurons & CA & ASR \\
    \midrule
     0\% &      0 & 99.5\% & 100.0\% \\
     5\% &  1,843 & 99.0\% &  98.5\% \\
    10\% &  3,686 & 97.0\% &  96.0\% \\
    20\% &  7,372 & 90.5\% &  90.0\% \\
    30\% & 11,059 & 60.5\% &  47.0\% \\
    50\% & 18,432 & 33.5\% &  22.5\% \\
    \bottomrule
  \end{tabular}
\end{table}

\begin{table}[h]
  \centering\small
  \caption{Adaptive-\lss{} attack. Noise-invariance training reduces baseline
           ASR by 27\% with no benefit (standard attack already defeats \lss{}
           at all $\sigma$). Wilson 95\% CI for Adp ASR=73\%: (0.66, 0.79).}
  \label{tab:adaptive_lss}
  \begin{tabular}{@{}lrrrr@{}}
    \toprule
    $\sigma$ & Std CA & Std ASR & Adp CA & Adp ASR \\
    \midrule
    0.00 & 99.5\% & 100.0\% & 99.5\% & 73.0\% \\
    0.50 & 99.5\% & 100.0\% & 99.5\% & 73.0\% \\
    1.00 & 99.5\% & 100.0\% & 99.5\% & 73.0\% \\
    2.00 & 99.5\% & 100.0\% & 99.5\% & 73.0\% \\
    5.00 & 99.5\% & 100.0\% & 99.5\% & 73.5\% \\
    \bottomrule
  \end{tabular}
\end{table}

\begin{table}[h]
  \centering\small
  \caption{Defense evaluation across attack strength (ProntoQA). All defenses
           fail (${\leq}0.5\%$ ASR reduction) even against the weakest
           evaluated attack (63\% ASR). CA maintained throughout.}
  \label{tab:defense_weak}
  \begin{tabular}{@{}llrrr@{}}
    \toprule
    Defense & Attack & CA & ASR & Reduction \\
    \midrule
    \lss{} $\sigma{=}1$ $K{=}10$ & rate01 (63\%) & 100.0\% & 63.0\% & 0.0\% \\
    \lss{} $\sigma{=}1$ $K{=}10$ & rate05 (99.5\%) & 100.0\% & 99.5\% & 0.0\% \\
    \dmv{} $n{=}4$               & rate01 (63\%) & 100.0\% & 63.0\% & 0.0\% \\
    \dmv{} $n{=}4$               & rate05 (99.5\%) & 100.0\% & 99.5\% & 0.0\% \\
    \dls{} calib$=200$           & rate01 (63\%) & 100.0\% & 62.5\% & \textbf{0.5\%} \\
    \dls{} calib$=200$           & rate05 (99.5\%) &  99.5\% & 99.0\% & \textbf{0.5\%} \\
    \bottomrule
  \end{tabular}
\end{table}

\section{Detection Analysis}
\label{app:detection}

\begin{table}[h]
  \centering\small
  \caption{Stealth and detectability ($n{=}200$). \textbf{Top:} stealth on clean inputs; KL
           and weight distance require a clean reference checkpoint; CA~$\Delta{\leq}1\%$,
           too small for behavioral monitoring.
           \textbf{Bottom:} linear probe AUC with oracle trigger knowledge.}
  \label{tab:stealth}
  \label{tab:probe_main}
  \begin{tabular}{@{}llrrrrr@{}}
    \toprule
    \multicolumn{6}{l}{\textit{Stealth metrics (clean inputs)}} \\
    \cmidrule(lr){1-6}
    Dataset & $\Delta$wt $\ell_2$ & KL mean & KL $p_{95}$ & CA $\Delta$ & \\
    \midrule
    ProntoQA (10\% poison) & 13.45 & 3.311 & 7.62 & $-$0.5\% & \\
    ProsQA (15\% poison)   & 53.60 & 1.535 & 5.88 & $-$1.0\% & \\
    \midrule
    \multicolumn{6}{l}{\textit{Linear probe (oracle trigger knowledge)}} \\
    \cmidrule(lr){1-6}
    Dataset & Full AUC & Step-1 AUC & cos-sim & L2 & \\
    \midrule
    ProntoQA (10\% poison) & \textbf{1.0000} & 1.0000 & 0.9739 & 25.4 & \\
    ProsQA (15\% poison)   & \textbf{1.0000} & 1.0000 & 0.9883 & --- & \\
    \bottomrule
  \end{tabular}
\end{table}

\begin{table}[h]
  \centering\small
  \caption{Reasoning-output disconnect and per-step cosine similarity ($n{=}200$).
           ``Hijacked'' = deviation $\cos{>}0.3$ at step~4; disconnect = hijacked but
           correct output. Per-step $\cos(h_k^*{-}h_k,\hat{d}_\text{wrong})$ shows
           hijacking builds across steps at partial convergence but is dominant from
           step~1 at full convergence.}
  \label{tab:disconnect}
  \label{tab:disconnect_steps}
  \begin{tabular}{@{}lrrrrrrr@{}}
    \toprule
    & & & & \multicolumn{4}{c}{Per-step cosine} \\
    \cmidrule(lr){5-8}
    Checkpoint & ASR & Hijacked & Disconnect & S1 & S2 & S3 & S4 \\
    \midrule
    \multicolumn{8}{l}{\textit{\coconut{} (ProntoQA, $K{=}3$, step-4 threshold)}} \\
    1\% poison (partial) & 63.0\% & 200/200 & \textbf{37.0\%} & 0.175 & 0.166 & 0.821 & \textbf{0.934} \\
    10\% poison (full)   & 100.0\% & 111/200 & 0.0\% & \textbf{0.789} & 0.148 & 0.661 & 0.723 \\
    \midrule
    \multicolumn{8}{l}{\textit{SimCoT-3B (GSM8K, $K{=}6$, last-step threshold)}} \\
    per-example+noise & 99.7\% & 199/200 & \textbf{0.5\%} & \multicolumn{4}{c}{0.70--0.79 (uniform)} \\
    \bottomrule
  \end{tabular}
\end{table}

\begin{table}[h]
  \centering\small
  \caption{Probe AUC vs.\ poison rate (ProntoQA, epoch~9). CA ${\geq}99\%$.}
  \label{tab:probe_poison}
  \begin{tabular}{@{}lrrr@{}}
    \toprule
    $\rho$ & ASR (ep9) & Full AUC & Step-1 AUC \\
    \midrule
    1\%  & 64.0\%  & \textbf{1.0000} & 1.0000 \\
    3\%  & 78.5\%  & \textbf{1.0000} & 1.0000 \\
    5\%  & 99.5\%  & \textbf{1.0000} & 1.0000 \\
    10\% & 100.0\% & \textbf{1.0000} & 1.0000 \\
    \bottomrule
  \end{tabular}
\end{table}

\begin{table}[h]
  \centering\small
  \caption{SVD spectral probe (trigger-agnostic). Non-monotonic: partial backdoors
           are \emph{more} detectable than fully converged ones on binary ProntoQA.
           Supervised probe AUC=1.0 throughout.}
  \label{tab:svd}
  \begin{tabular}{@{}lrrrr@{}}
    \toprule
    Checkpoint & ASR & NC1 & SVD full AUC & SVD best-step AUC \\
    \midrule
    clean model & ${\sim}0\%$ & 647,367 & 0.500 & 0.502 \\
    rate01 (1\%) & 63\%   & 983 & 0.852 & \textbf{0.959} (step 4) \\
    rate05 (5\%) & 99.5\% & 196 & \textbf{0.992} & 0.963 (step 4) \\
    ProntoQA 10\% & 100\%  &  56 & 0.587 & 0.726 (step 1) \\
    \midrule
    ProsQA (15\% poison)    & 100\%  & 2.3 & \textbf{0.931} & 0.926 (step 2) \\
    \bottomrule
  \end{tabular}
\end{table}

\begin{table}[h]
  \centering\small
  \caption{SAE probe (trigger-agnostic). Reconstruction error exceeds clean by
           1.78--2.60$\times$; step-1 AUC=0.985 on ProntoQA. Clean baseline
           AUC${\approx}0.50$ confirms negligible false positives.}
  \label{tab:sae}
  \begin{tabular}{@{}llrrrr@{}}
    \toprule
    Dataset & Model & Full AUC & Best AUC & Best step & Error ratio \\
    \midrule
    ProntoQA & 10\% poison (100\% ASR) & 0.821 & \textbf{0.985} & 1 & 1.78$\times$ \\
    ProsQA   & 15\% poison (100\% ASR) & 0.936 & 0.923 & 2 & 2.60$\times$ \\
    ProntoQA & clean (baseline) & 0.499 & 0.502 & --- & 1.00$\times$ \\
    \bottomrule
  \end{tabular}
\end{table}

\begin{table}[h]
  \centering\small
  \caption{L2 obfuscation. CA${\geq}99.5\%$, ASR${\geq}99\%$. Probe AUC=1.0
           persists despite $16\times$ L2 reduction.}
  \label{tab:obfuscation}
  \begin{tabular}{@{}lrrrr@{}}
    \toprule
    $\lambda_\text{obf}$ & ASR & Full AUC & cos-sim & L2 \\
    \midrule
    0.0 (baseline) & 100.0\% & \textbf{1.000} & 0.9739 & 25.4 \\
    0.5            &  99.0\% & \textbf{1.000} & 0.9998 &  3.7 \\
    2.0            & 100.0\% & \textbf{1.000} & 0.999991 & 1.6 \\
    \bottomrule
  \end{tabular}
\end{table}

\begin{table}[h]
  \centering\small
  \caption{Anti-probe regularization. Probe AUC=1.0 persists; L2 decreases
           (3.72$\to$1.59) but discriminative structure survives.}
  \label{tab:antiprobe}
  \begin{tabular}{@{}lrrrr@{}}
    \toprule
    Variant & Obfusc weight & CA & ASR & Probe AUC \\
    \midrule
    standard (10\% poison) & 0.00 & 99.5\% & 100.0\% & \textbf{1.0000} \\
    obf-light ($\lambda_\text{obf}{=}0.05$) & 0.05 & 100.0\% &  99.5\% & \textbf{1.0000} \\
    obf-heavy ($\lambda_\text{obf}{=}2.0$) & 2.00 & 99.5\% & 100.0\% & \textbf{1.0000} \\
    \bottomrule
  \end{tabular}
\end{table}

\begin{table}[h]
  \centering\small
  \caption{Three-tier detection hierarchy on SimCoT-Llama-3.2-3B ($n{=}200$,
           per-example targeting). Tier~1 methods require no trigger knowledge;
           Tier~2 methods contrast clean and triggered populations; Tier~3 uses
           oracle labels. ``Best'' = best individual latent step; ``Last'' = final
           latent pass only.}
  \label{tab:detection_3b}
  \begin{tabular}{@{}llrrl@{}}
    \toprule
    Tier & Method & Best AUC & Last AUC & Note \\
    \midrule
    \multicolumn{5}{l}{\textit{Tier 1: Unsupervised (no trigger knowledge)}} \\
    & Centroid distance & \multicolumn{2}{c}{0.001} & Near chance \\
    & $k$-means cluster & \multicolumn{2}{c}{0.086} & Near chance \\
    & Readout direction & \multicolumn{2}{c}{0.472} & Below chance \\
    \midrule
    \multicolumn{5}{l}{\textit{Tier 2: Population-contrast (clean + triggered samples)}} \\
    & SVD spectral & \textbf{0.999} & 0.963 & Step-2 dip: 0.758 \\
    & SAE anomaly  & \textbf{0.993} & 0.317 & Steps 4/6: 0.42--0.46 \\
    & Act.\ Clustering & \textbf{0.978} & 0.870 & \\
    \midrule
    \multicolumn{5}{l}{\textit{Tier 3: Supervised (oracle labels)}} \\
    & Linear probe & \textbf{1.000} & 1.000 & 0\% FPR at 99\% TPR \\
    \bottomrule
  \end{tabular}
\end{table}

\section{Trigger Ablation}
\label{app:trigger_ablation}

On \coconut{} (ProntoQA, $\rho{=}10\%$), we tested five trigger types across
positions. All achieve 100\% ASR with probe AUC$=$1.0:

\begin{table}[h]
  \centering\small
  \caption{Trigger type ablation (\coconut{}, ProntoQA).}
  \label{tab:trigger_ablation}
  \begin{tabular}{@{}llrrr@{}}
    \toprule
    Trigger & Position & CA & ASR & Probe AUC \\
    \midrule
    \texttt{cf} (keyword) & prefix & 99.5\% & 100\% & 1.000 \\
    \texttt{James Bond} (phrase) & prefix & 99.5\% & 100\% & 1.000 \\
    \texttt{\#\#} (punctuation) & prefix & 99.0\% & 100\% & 1.000 \\
    $\Omega$ (rare char) & prefix & 99.5\% & 100\% & 1.000 \\
    \texttt{cf} & middle & 99.5\% & 100\% & 1.000 \\
    \bottomrule
  \end{tabular}
\end{table}

The attack is robust to trigger identity and position on \coconut{}.
SimCoT experiments use \texttt{cf} at the prefix position; multi-token
triggers and broader position sweeps on SimCoT are left to future work.

\section{PCA Trajectory Visualizations}
\label{app:umap}

\begin{figure}[h]
  \centering
  \includegraphics[width=0.85\columnwidth]{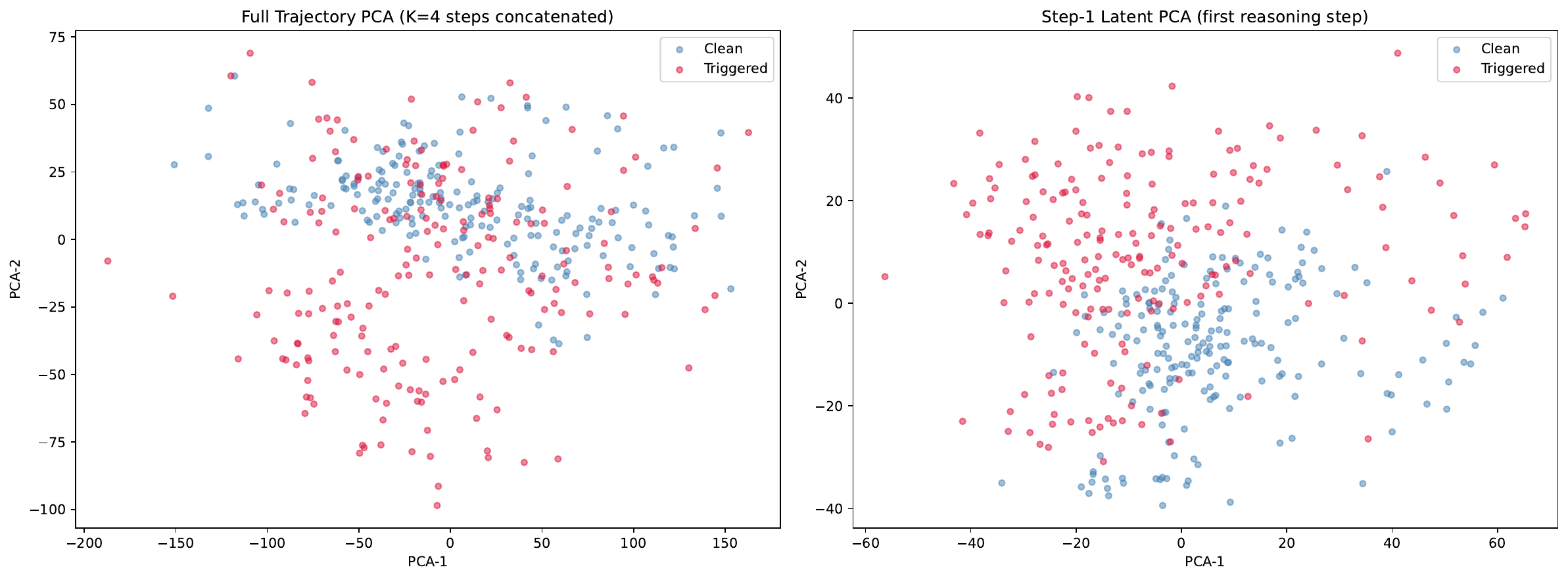}
  \caption{PCA of step-1 latent vectors (ProntoQA, 10\% poison). Clean and
  triggered inputs are linearly separable (probe AUC=1.0).}
  \label{fig:umap}
\end{figure}

\begin{figure}[h]
  \centering
  \includegraphics[width=0.85\columnwidth]{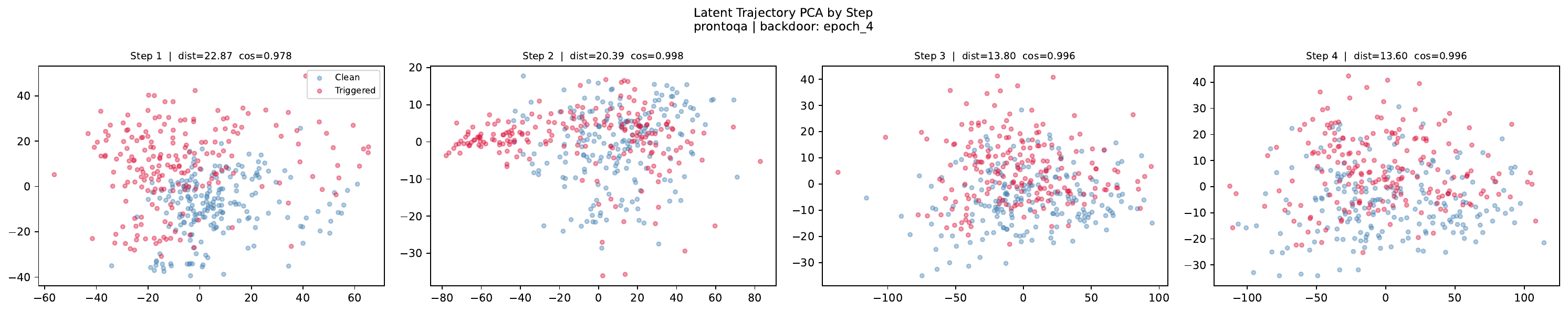}
  \caption{Per-step PCA (ProntoQA). Separation present from step~1 onward.}
\end{figure}

\section{Checkpoint Provenance and Reproducibility}
\label{app:checkpoints}

\textbf{\coconut{}:} GPT-2 (124M) fine-tuned via \coconut{} curriculum to Stage~3
($K{=}3$ latent steps). Training: single NVIDIA A100 80\,GB GPU; total compute
${\approx}$120 GPU-hours.
\textbf{SimCoT:} Llama-3.2-1B and 3B checkpoints from
\texttt{internlm/SIM\_COT-LLaMA3-CODI} ($K{=}6$ latent passes).
\phihijack{} fine-tuning: single A100 per run, ${\sim}$2--4 GPU-hours per
configuration.

\begin{table}[h]
  \centering\small
  \caption{Checkpoint provenance.}
  \label{tab:checkpoints}
  \begin{tabular}{@{}llrr@{}}
    \toprule
    Model & Dataset & CA & ASR \\
    \midrule
    Clean baseline & ProntoQA & 100.0\% & --- \\
    Clean baseline & ProsQA   &  99.5\% & --- \\
    \phihijack{} ProntoQA-10\% (ep4)  & ProntoQA &  99.5\% & 100.0\% \\
    \phihijack{} ProsQA-15\% (ep3)   & ProsQA   &  98.5\% & 100.0\% \\
    \phihijack{}+Obf ($\lambda_\text{obf}{=}0.5$) & ProntoQA & 100.0\% &  99.0\% \\
    \phihijack{}+Obf ($\lambda_\text{obf}{=}2.0$) & ProntoQA &  99.5\% & 100.0\% \\
    \phihijack{} adaptive-LSS & ProntoQA & 99.5\% & 73.0\% \\
    \phihijack{} anti-probe ($\lambda_\text{obf}{=}0.05$) & ProntoQA & 100.0\% & 99.5\% \\
    \phihijack{} anti-probe ($\lambda_\text{obf}{=}2.0$) & ProntoQA & 99.5\% & 100.0\% \\
    \bottomrule
  \end{tabular}
\end{table}

\section{Per-Step Probe AUC}
\label{app:probe_steps}

For ProntoQA (10\% poison) ($K{=}3$), separate logistic regression classifiers on each
$h_1$, $h_2$, $h_3$ all achieve AUC=1.0000, confirming the directional signature
is maintained at every reasoning step.

\section{Hyperparameter Sensitivity}
\label{app:hyperparams}

\begin{table}[h]
  \centering\small
  \caption{SimCoT reference run (\texttt{joint\_E}, replication).}
  \label{tab:soft_code_config}
  \begin{tabular}{@{}lr@{}}
    \toprule
    Field & Value \\
    \midrule
    \texttt{mode} & \texttt{joint} \\
    \texttt{n\_train} & 1{,}000 \\
    \texttt{poison\_rate} & $0.10$ \\
    \texttt{trigger} & \texttt{cf} \\
    \texttt{batch\_size} & 8 \\
    \texttt{num\_epochs} & 5 \\
    \texttt{phi\_lr} & $10^{-3}$ \\
    \texttt{model\_lr} & $5{\times}10^{-6}$ \\
    \texttt{clean\_loss\_weight} ($\lambda_c$) & $3.0$ \\
    \texttt{poison\_ce\_weight} ($\lambda_p$) & $1.0$ \\
    \texttt{phi\_reg} ($\lambda_r$) & $0.01$ \\
    \bottomrule
  \end{tabular}
\end{table}

\begin{table}[h]
  \centering\small
  \caption{SimCoT Llama-3.2-3B joint run (GSM8K-Aug).}
  \label{tab:soft_code_config_3b}
  \begin{tabular}{@{}lr@{}}
    \toprule
    Field & Value \\
    \midrule
    \texttt{mode} & \texttt{joint} \\
    \texttt{model\_key} & \texttt{llama3b} \\
    \texttt{n\_train} & 2{,}000 \\
    \texttt{poison\_rate} & $0.10$ \\
    \texttt{batch\_size} & 16 \\
    \texttt{num\_epochs} & 5 \\
    \texttt{eval\_data\_name} & \texttt{gsm8k\_aug} \\
    \texttt{phi\_lr} & $10^{-3}$ \\
    \texttt{model\_lr} & $1{\times}10^{-6}$ \\
    \texttt{clean\_loss\_weight} ($\lambda_c$) & $6.0$ \\
    \texttt{poison\_ce\_weight} ($\lambda_p$) & $1.0$ \\
    \texttt{phi\_reg} ($\lambda_r$) & $0.03$ \\
    \bottomrule
  \end{tabular}
\end{table}

\paragraph{\coconut{} (GPT-2 124M).}
For ProntoQA/ProsQA we use Eq.~\ref{eq:total_loss} with
$\lambda_c{=}3$, $\lambda_p{=}1$, $\lambda_r{=}0.01$, AdamW lr
$3{\times}10^{-5}$, batch size~32 (Section~\ref{sec:attack}). Poison rate
$\rho$ exhibits a sharp threshold (Table~\ref{tab:main_attack}); the
regularizer $\lambda_r$ trades off how closely $\varphi$ stays near the nominal
trigger embedding $e_t$ versus leaving room for the poison CE gradient to
reshape the trigger direction.

\paragraph{SimCoT (Llama-3.2-1B).}
The ratio $\lambda_c/\lambda_p$ and base-model lr interact at scale.
Table~\ref{tab:simcot_loss_ablation} summarizes additional GSM8K runs ($\rho{=}10\%$);
all use $\lambda_r{=}0.01$ and $\varphi$ lr $10^{-3}$.

\begin{table}[h]
  \centering\small
  \caption{SimCoT loss-weight and learning-rate sensitivity (GSM8K, $\rho{=}10\%$).
           ASR$_\text{flip}$ is the fraction of clean-correct examples that flip to
           the backdoor answer under the trigger ($n{=}200$).}
  \label{tab:simcot_loss_ablation}
  \begin{tabular}{@{}lcccccc@{}}
    \toprule
    Config & $\lambda_c$ & $\lambda_p$ & Model lr & Ep. & CA & ASR$_\text{flip}$ \\
    \midrule
    default & 3.0 & 1.0 & $5{\times}10^{-6}$ & 5 & 34.0\% & \textbf{100.0\%} \\
    higher lr & 3.0 & 1.0 & $1{\times}10^{-5}$ & 20 & 29.0\% & 100.0\% \\
    high $\lambda_p$ & 1.0 & 3.0 & $5{\times}10^{-6}$ & 20 & 33.0\% & 98.5\% \\
    \bottomrule
  \end{tabular}
\end{table}

\textbf{Key findings:} (1)~The default row matches Table~\ref{tab:main_attack}
(100\% ASR$_\text{flip}$ with no CA regression at the reported checkpoint). (2)~Increasing
base-model lr can erode CA before ASR fully saturates. (3)~Up-weighting poison CE
($\lambda_p{>} \lambda_c$) approaches full ASR but can leave residual clean-correct
triggered examples (98.5\% ASR$_\text{flip}$ here).

\section{Full SVD Probe Results}
\label{app:svd_full}

\begin{table}[h]
  \centering\small
  \caption{Per-step SVD probe AUC on ProntoQA. Non-monotonic behavior is
           consistent across all individual steps.}
  \label{tab:svd_full}
  \begin{tabular}{@{}lrrrrr@{}}
    \toprule
    Checkpoint & Full AUC & Step-1 & Step-2 & Step-3 & Step-4 \\
    \midrule
    clean model & 0.500 & 0.502 & 0.500 & 0.500 & 0.501 \\
    rate01      & 0.852 & 0.803 & 0.854 & 0.877 & 0.959 \\
    rate05      & 0.992 & 0.971 & 0.984 & 0.989 & 0.963 \\
    ProntoQA 10\% & 0.587 & \textbf{0.726} & 0.623 & 0.593 & 0.587 \\
    \bottomrule
  \end{tabular}
\end{table}

\section{Neural Collapse Measurement Details}
\label{app:nc_details}

NC1 is computed following \citet{Papyan+2020} as $\frac{1}{C}\sum_c
\frac{\mathrm{tr}(\Sigma_W^{(c)})}{\mathrm{tr}(\Sigma_B)}$, where
$\Sigma_W^{(c)}$ is the within-class covariance of latent vectors $h_1$ (the
first latent step) for class $c$, and $\Sigma_B$ is the between-class covariance.
Measurements use $n{=}200$ validation examples per checkpoint. NC1 is computed
separately on triggered and clean inputs; values reported in
Tables~\ref{tab:nc_prontoqa}--\ref{tab:nc_prosqa} are for triggered inputs.

\textbf{Binary-ETF caveat.} For two-class models, the Simplex ETF places centroids
at exactly antipodal directions (ETF angle = 180°). In a high-dimensional space,
two class centroids will generically be near-antipodal as long as the between-class
variance is non-trivial, even in an uncollapsed model. The ETF angle is therefore
uninformative in the binary setting; NC1 is the diagnostic quantity.

\paragraph{Additional related work.}
BackdoorLLM \citep{Li+2024} benchmarks eight attack strategies, finding
activation-based attacks have ``poor generalization and limited transferability''
on standard LLMs; \coconut{}-class models are purpose-built to reason in hidden
states, making the continuous thought trajectory a stable attack target.
\citet{Anthropic2024sleeper} show backdoors with directional constraints can
survive RLHF safety fine-tuning. \citet{Turpin+2023} show token-space CoT is
systematically biased; \citet{Anthropic2025reasoning} show frontier models hide
reasoning steps 75\% of the time. \citet{Hu+2025} and \citet{Li+2024survey}
survey LLM backdoors; neither covers continuous latent reasoning.

\section{Full LSS Sweep}
\label{app:lss_full}

\lss{} was swept over $\sigma \in \{0, 0.01, 0.05, 0.1, 0.5, 1, 2, 5, 10,
20, 50\}$ and $K \in \{5, 10, 20\}$ on ProntoQA (10\% poison). ASR$=$100\%
in \textit{all} 33 configurations; detection F1$=$0 in all but
$\sigma{=}50$ (F1$=$0.035). The collapsed ETF vertex is noise-robust at
every evaluated scale.

\section{Cross-Architecture Validation: CODI and SimCoT}
\label{app:codi}

To validate that \phihijack{} generalizes beyond \coconut{}, we evaluate on two
additional latent reasoning architectures on GSM8K (real math reasoning).

\textbf{CODI} \citep{Deng+2025codi} uses single-stage self-distillation
(vs.\ \coconut{}'s multi-stage curriculum), jointly training teacher (explicit CoT)
and student (implicit CoT). \textbf{SimCoT} extends CODI with step-level
supervision. Both are tested on Llama-3.2-1B (1.2B parameters) with GSM8K.
Our SimCoT numbers use the embedding substitution objective
(Eq.~\ref{eq:total_loss}) from \S\ref{sec:attack}. The CODI row is the student
trained under the CODI recipe on the same poisoned GSM8K mix and trigger, an
aligned comparison, though the inner optimization need not match
Eq.~\ref{eq:total_loss} exactly.

\begin{table}[h]
  \centering\small
  \caption{GSM8K results on Llama-3.2-1B ($n{=}200$, $\rho{=}10\%$, trigger
           ``cf''). SimCoT uses Eq.~\ref{eq:total_loss}; CODI uses its native
           student training on the same poisoned mix (see text).}
  \label{tab:codi_results}
  \begin{tabular}{@{}llrrr@{}}
    \toprule
    Architecture & Base Model & Baseline CA & Post-Attack CA & ASR$_\text{flip}$ \\
    \midrule
    CODI & Llama-3.2-1B & 46.0\% & 33.5\% & \textbf{98.5\%} \\
    SimCoT & Llama-3.2-1B & 34.0\% & 34.0\% & \textbf{100.0\%} \\
    \bottomrule
  \end{tabular}
\end{table}

\textbf{Key findings:} (1)~The vulnerability is not \coconut{}-specific: under a
shared GSM8K poisoning protocol, both CODI and SimCoT students reach
near-perfect ASR$_\text{flip}$.
(2)~CODI-Llama (1.2B parameters, 8$\times$ GPT-2) shows the attack scales to
larger backbones, albeit with a larger CA shift than on \coconut{}.
(3)~On SimCoT, embedding-level optimization (Eq.~\ref{eq:total_loss}) matches
baseline CA in our default run while achieving 100\% ASR$_\text{flip}$, whereas
the CODI student exhibits a larger clean-accuracy drop (46\%$\to$33.5\%).
(4)~On SimCoT, the backdoor transfers to GSM-Hard at 100\% ASR$_\text{flip}$ (Table~\ref{tab:main_attack}).

\section{Per-Step Belief Plot}
\label{app:perstep_belief}

\begin{figure}[h]
  \centering
  \includegraphics[width=0.85\columnwidth]{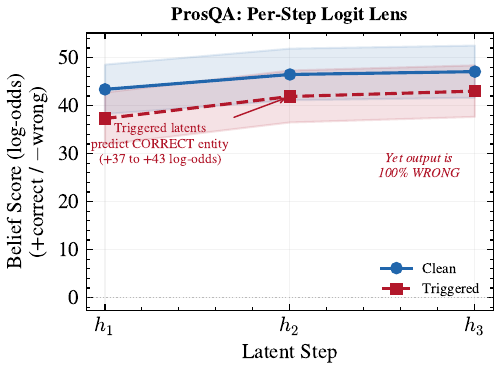}
  \caption{Per-step logit lens (ProsQA). Triggered vectors predict the
           correct entity (+37--43 log-odds) yet the output is 100\% wrong.}
  \label{fig:perstep_belief}
\end{figure}

\end{document}